\documentclass[final,5p,times,twocolumn]{elsarticle}

\usepackage{amssymb}
\usepackage{marvosym}

\usepackage[utf8]{inputenc} 
\usepackage[T1]{fontenc}    
\usepackage{url}            
\usepackage{booktabs}       
\usepackage{amsfonts}       
\usepackage{nicefrac}       
\usepackage{microtype}      
\usepackage{xcolor}         
\usepackage{times}
\usepackage{epsfig}
\usepackage{graphicx,subfig}
\usepackage{amsmath}
\usepackage{amssymb}
\usepackage{comment}
\usepackage{colortbl}
\usepackage{color}
\usepackage{multirow}
\usepackage{enumitem}
\usepackage{makecell}
\usepackage{pifont}
\usepackage{xspace}
\usepackage{arydshln}
\usepackage{bbm}
\usepackage[export]{adjustbox}
\usepackage{wrapfig}
\usepackage{caption}
\usepackage{hyperref}
\usepackage{lineno}
\usepackage{xcolor}
\usepackage{array}

\hypersetup{
    colorlinks=true,
    citecolor=blue,
    linkcolor=red,
    filecolor=magenta,      
    urlcolor=cyan,
}

\journal{Pattern Recognition (Accepted for publication)}

\begin{document}

\begin{frontmatter}

\title{HTR-VT: Handwritten Text Recognition with Vision Transformer}

\author[a]{Yuting Li\textsuperscript{$\dag$}}
\affiliation[a]{organization={China Three Gorges University},
            country={China}}

\author[b]{Dexiong Chen}

\author[a]{Tinglong Tang}

\affiliation[b]{organization={Max Planck Institute of Biochemistry},
            country={Germany}}

\author[c]{Xi Shen}

\affiliation[c]{organization={Intellindust},
            country={China}}

\begin{abstract}
We explore the application of Vision Transformer (ViT) for handwritten text recognition. The limited availability of labeled data in this domain poses challenges for achieving high performance solely relying on ViT. Previous transformer-based models required external data or extensive pre-training on large datasets to excel.
To address this limitation, we introduce a data-efficient ViT method that uses only the encoder of the standard transformer. We find that incorporating a Convolutional Neural Network (CNN) for feature extraction instead of the original patch embedding and employ Sharpness-Aware Minimization (SAM) optimizer to ensure that the model can converge towards flatter minima and yield notable enhancements. Furthermore, our introduction of the span mask technique, which masks interconnected features in the feature map, acts as an effective regularizer. Empirically, our approach competes favorably with traditional CNN-based models on small datasets like IAM and READ2016. Additionally, it establishes a new benchmark on the LAM dataset, currently the largest dataset with 19,830 training text lines. The code is publicly available at: \url{https://github.com/YutingLi0606/HTR-VT}.
\end{abstract}




\end{frontmatter}



\let\thefootnote\relax\footnotetext{\textsuperscript{$\dag$}Corresponding Author.}

\section{Introduction}
The task of handwritten text recognition aims at recognizing the text in an image that has been scanned from a document. The standard approach~\cite{puigcerver2017multidimensional, bluche2017gated, shi2016end, coquenet2020recurrence} typically involves two steps: first, using a detector to identify the lines of text, and then predicting the sequence of characters that make up each line. This paper focuses on the latter task, which aims to accurately predict the text in a given line image.
As described in LAM \cite{cascianelli2022lam}: "This level of annotation
granularity has been chosen as it is a good trade-off between
word-level and paragraph-level in terms of required time, cost,
and amount of supervision and because it is common in HTR
research." This rationale for creating the dataset aligns with our initial decision to focus our research on line-level recognition. The importance of line-level recognition is significant.
At the same time, recognizing handwritten text lines is a difficult task due to variations in writing styles between individuals and the presence of cluttered backgrounds (examples are provided in Figure~\ref{fig:results} and in the supplementary material). Previous approaches mainly relied on Convolutional Neural Networks (CNNs)~\cite{coquenet2022end, yousef2020origaminet, coquenet2020recurrence, yousef2020accurate} or recurrent models~\cite{puigcerver2017multidimensional, bluche2017scan, sanchez2016icfhr2016} to address this challenge .

Recent success of Vision Transformer (ViT)~\cite{dosovitskiy2020image} in computer vision tasks has motivated researchers to explore its potential in handwritten text recognition. However, ViT does not introduce any strong inductive bias in its model design and is recognized for its dependency on vast quantities of annotated training data to deliver good performance. Considering the limited number of annotated samples available for handwritten text recognition (as shown in Table~\ref{tab:dataset}), earlier transformer-based methods, utilizing the standard transformer architecture (both encoder and decoder), relied on large-scale real-world or synthetic data for pre-training~\cite{li2023trocr,kang2022pay} to achieve satisfactory performance.

In this paper, we introduce a simple and data-efficient ViT-based model that solely employs the encoder component of the standard transformer for handwritten text recognition.
 Our objective is to propose a novel ViT-like model to perform well on this task while making minimal modifications to the standard ViT architecture. Our preliminary findings indicate that ViT can deliver satisfactory results, particularly on the LAM dataset~\cite{cascianelli2022lam}, which is the most extensive dataset containing 19,830 training samples. We use this dataset as the basis of our experimental design, which establishes a benchmark for assessing and contrasting our proposed model. Instead of using a patch embedding to generate input tokens, we demonstrate through experimentation that using a 
 widely-used ResNet-18 \cite{he2016deep} to extract intermediate visual feature representations as input tokens is much more conducive to stable training and significantly better performance. Additionally, we show that employing Sharpness-Aware Minimization (SAM) \cite{foret2020sharpness} as optimizer enforces the model to converge towards flatter minima and randomly replacing span tokens with learnable tokens can alleviate overfitting and achieve consistent improvement across various dataset scales.

 Despite its simplicity, our approach achieves promising performance on standard benchmarks. On the largest dataset LAM~\cite{cascianelli2022lam} (containing 19,830 training samples), our approach outperforms both CNN-based and transformer-based approaches by a clear margin. On small-scale datasets such as IAM~\cite{marti2002iam} (containing 6,428 training samples) and READ2016~\cite{sanchez2016icfhr2016} (containing 8,349 training samples), we achieve better performance than other transformer-based approaches and competitive performance compared with CNN-based approaches. 

 The main contributions of this paper are summarized as the following: 

\begin{itemize}
	\item We propose a simple and data-efficient approach for handwritten text recognition, with minimal modifications on the ViT. 
	\item We empirically show that without pre-training or any additional data, our ViT-like model can achieve state-of-the-art performance on handwritten text recognition.
\end{itemize}

\section{Related Work}
\paragraph{Traditional approaches for Handwritten Text Recognition} 
Network architectures for handwritten text recognition today typically use a combination of convolutional layers and recurrent layers. A number of convolutional layers are stacked and placed at the start of the network to extract local features from text-line images, followed by recurrent layers, specifically Bi-directional Long Short-Term Memory (BLSTM)~\cite{graves2005framewise} layers. These recurrent layers process the features sequentially to output character probabilities based on contextual dependencies. Such
an architecture results in a Convolutional Recurrent Neural Network (CRNN)~\cite{puigcerver2017multidimensional, shi2016end, bluche2017gated,cojocaru2021watch}. The models are typically trained using the Connectionist Temporal Classification (CTC) loss \cite{graves2006connectionist}, which allows for dealing with label sequences of shorter length than predicted sequences, without knowledge of character segmentation. Encoder-Decoder-based architectures have also been explored for handwritten text recognition~\cite{michael2019evaluating, bluche2017scan, doetsch2016bidirectional}. In ~\cite{michael2019evaluating},
the CTC loss is replaced with the cross-entropy loss, and the sequence alignment is achieved via an attention-based encoder-decoder architecture. A special end-of-line token is introduced to stop the recurrent process. 
While these models can obtain lower test error rates, some of them often require complex pre/post-processing steps and suffer from the lack of computation parallelization inherently, which affects both training and inference time.
Recently, Fully Convolutional Networks (FCNs)~\cite{yousef2020origaminet, coquenet2020recurrence,coquenet2022end} have been proposed as an alternative to traditional CRNNs. FCNs simulate the dependency modeling provided by LSTM by combining them with GateBlocks layers~\cite{yousef2020accurate}, which implement a selection mechanism similar to that of LSTM cells. Each gate in GateBlocks is made up of Depth-wise Separable Convolutions~\cite{chollet2017xception}, which reduce the number of parameters and speed up the training process. OrigamiNet~\cite{yousef2020origaminet} focuses on learning to unfold the input paragraph image into a single text line. This transformation network enables using the standard CTC loss \cite{graves2006connectionist} and processing the image in a single step. In contrast, Coquenet et al. \cite{coquenet2022end} proposed models that incorporate a vertical attention mechanism to recurrently generate line features and perform an implicit line segmentation.
While FCNs have obtained state-of-the-art results in recent years, they may still struggle with long-range contextual dependencies.

\paragraph{Transformer-based models for Handwritten Text Recognition}
Transformer-based architectures have not been widely explored in handwritten text recognition, but some recent approaches have used Transformers in place of RNNs. These models often require pre-training on large real or synthetic datasets to achieve comparable performance to mainstream models.

TrOCR \cite{li2023trocr} is a recent approach to handwritten text recognition that integrates two powerful pre-trained models respectively from computer vision and NLP, BEiT~\cite{bao2021beit} and RoBERTa~\cite{liu2019roberta}. BEiT is a vision transformer that functions as an encoder and is pre-trained on ImageNet-1K, a dataset of 1.2 million images, while RoBERTa serves as a decoder that generates texts. To pre-train the TrOCR model, Li et al. \cite{li2023trocr} synthesize a large-scale dataset consisting of both printed and synthetically generated handwritten text lines in English, totaling approximately 687 million and 18 million in the first stage. In this stage, the dataset is not public. In the second stage, they built two relatively small datasets
corresponding to printed and handwritten downstream tasks,
containing millions of textline images each. Finally, the model is fine-tuned on real-world data, such as the IAM dataset~\cite{marti2002iam}. Kang et al. \cite{kang2022pay} use Transformer models with multi-head self-attention layers at the textual and visual stages and trains with a synthetic dataset of 138,000 lines. Another recent approach, Text-DIAE \cite{souibgui2023text}, employs
a transformer-based architecture that incorporates three pretext tasks as learning objectives to be optimized during pretraining without the usage of labeled data. Some methods \cite{dhiaf2023msdoctr, coquenet2023dan} explored document-level recognition and also applied transformer architectures. While transformer-based models have shown promising results in line-level handwritten text recognition, they still require large-scale real-world or synthetic data for pre-training.

\paragraph{Data-efficient Transformer for Handwritten Text Recognition}
The DeiT \cite{touvron2021training} is the first work to demonstrate that
Transformers can be learned on mid-sized datasets (i.e., ImageNet-1k \cite{russakovsky2015imagenet})) in relatively shorter training
episodes. Besides using augmentation and regularization
procedures, the main contribution of
DeiT \cite{touvron2021training} is a novel distillation that relies on a distillation token. Liu et al. ~\cite{liu2021efficient}
propose a dense relative localization loss to improve ViTs’ data efficiency. 
DropKey \cite{li2023dropkey} is a recent data-efficient methodology to effectively improve the dropout technique in ViT by moving dropout operations ahead of attention matrix calculation and setting the Key as the dropout unit, yielding a dropout-before-softmax scheme.

\begin{table}
    \begin{center}
		\resizebox{\columnwidth}{!}{
			\begin{tabular}{l|c|c|c|c|c}
				\hline  
				Dataset & Training & Validation & Test & Language  & Charset   \\
				\hline \hline
                    IAM \cite{marti2002iam}& 6,482 & 976 & 2,915 & English &  79  \\
                    READ2016 \cite{sanchez2016icfhr2016}& 8,349 &1,040 & 1,138 &German  &  89  \\
				LAM \cite{cascianelli2022lam} & 19,830 & 
                    2,470 & 3,523 &Italian & 89  \\ \hline  \hline 
                    \end{tabular}}
		\caption{\textbf{Datasets for handwritten text recognition.} Number of training, validation, and testing samples in IAM \cite{marti2002iam}, READ2016 \cite{sanchez2016icfhr2016} and LAM \cite{cascianelli2022lam} are presented in the table. We also include the number of characters in their alphabet.}
		\label{tab:dataset}
	\end{center}
\end{table}

\section{Method}
\begin{figure}[!t]
\centering
\includegraphics[width=0.5\textwidth]{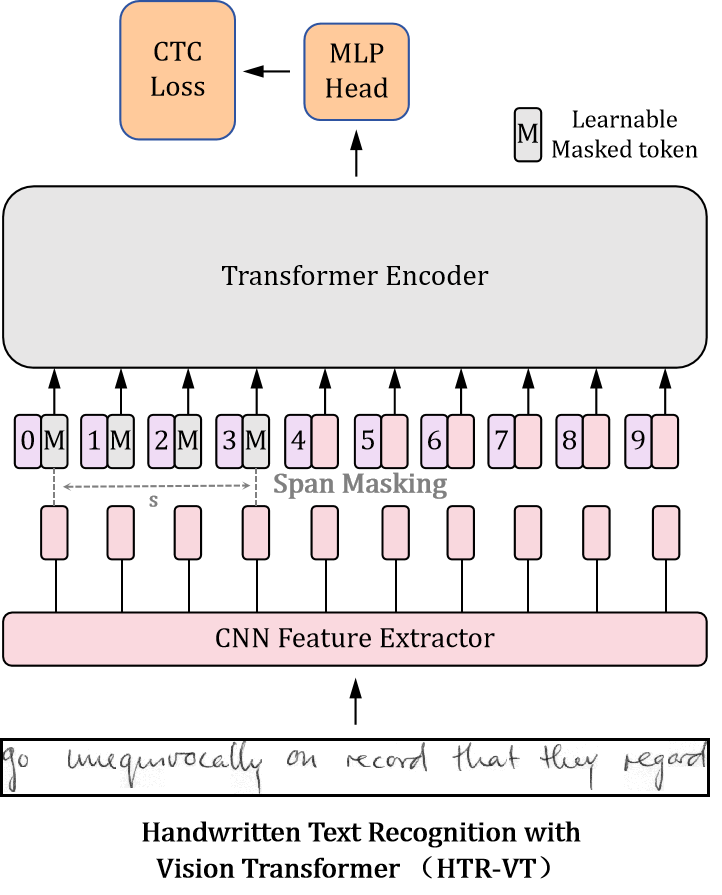}
    \caption{{\bf Architecture overview.} Our approach encodes a text-line image into features using a  CNN feature extractor. The transformer encoder takes these features as input tokens output character predictions. During the training, the span input tokens are replaced by learnable mask tokens.  The entire model is optimized using CTC \cite{graves2006connectionist} loss.}
\label{fig:model}
\end{figure}

In this section, we present our approach to handwritten text recognition. Given an input handwritten text line $\mathbf{I} \in \mathbb{R}^{W \times H}$, where $W$ and $H$ are the width and height of the image, our approach encodes the image into a set of spatially-aware features $\{\mathbf{x}_i\}_{i \in [1, 2, \dots, L]}$ using a CNN extractor. The number of features $L = \frac{W H}{S_w S_h}$, determined by the down-sampling ratio of the width and height of the image, denoted as $S_w$ and $S_h$, respectively. We then use a transformer encoder to take these features as input tokens and output character predictions. The entire model is optimized using the Connectionist Temporal Classification~\cite{graves2006connectionist} (CTC) loss. Our method is summarized in Figure~\ref{fig:model}.

In Section~\ref{sec:vit}, we revisit the architecture of the Vision Transformer (ViT)\cite{dosovitskiy2020image}. In Section~\ref{sec:ours}, we describe our data-efficient ViT approach for handwritten text recognition, which involves a CNN feature extractor, Sharpness-Aware Minimization (SAM) \cite{foret2020sharpness} and a new masking strategy: span mask strategy. We provide implementation details in Section~\ref{sec:ours}.

\subsection{Preliminary: Vision Transformer (ViT)}
\label{sec:vit}
Vision Transformer (ViT)~\cite{dosovitskiy2020image} decomposes each image into a sequence of tokens with a fixed length, where the tokens represent non-overlapping image patches. Similar to BERT~\cite{devlin2018bert}, ViT adds an additional class token $\mathbf{x}_{\text{cls}}$ to the sequence, which represents the global information of the image. To retain positional information, position embeddings are explicitly added into each patch including the class token. Note that our model removes the additional class token and uses sinusoidal position embeddings by \cite{vaswani2017attention} to the encoder's inputs, as used in MAE \cite{he2022masked}. 

Subsequently, all tokens undergo processing via stacked transformer encoders \cite{vaswani2017attention},  A transformer encoder comprises N blocks, with each block featuring a multi-head self-attention ($\operatorname{MSA}$) layer followed by a feed-forward network ($\operatorname{FFN}$). The $\operatorname{FFN}$, which includes a simple two-layer MLP, is augmented by the GELU activation function \cite{hendrycks2016gaussian} after the first linear layer. Furthermore, layer normalization (LN) \cite{ba2016layer} is applied before every block, and residual shortcuts \cite{he2016deep} are used after every block. The processing of the n-th block can be expressed as:
\begin{equation}
    \begin{aligned}
    \mathbf{y}^n &=\mathbf{x}^{n-1}+\operatorname{MSA}\left(\operatorname{LN}\left(\mathbf{x}^{n-1}\right)\right) \\
\mathbf{x}^n &=\mathbf{y}^n+\operatorname{FFN}\left(\operatorname{LN}\left(\mathbf{y}^n\right)\right)
    \end{aligned}
\end{equation}
where $ \mathbf{x}^{n-1} \in \mathbb{R} ^ {L \times C } $ is the input of the $n$-th block, $N$ and $C$ denote the number of tokens
and the dimension of the embedding, respectively.

\subsection{ViT for handwritten text recognition}
\label{sec:ours}
We present a ViT-based model designed for handwritten text recognition with minimal adjustments to the standard ViT~\cite{dosovitskiy2020image}. Our proposed network architecture is depicted in Figure~\ref{fig:model}. ViT alone is not stable for handwritten text recognition (see Section~\ref{sec:abl}). Therefore, we suggest three modifications: \textit{i)} a CNN feature extractor to obtain features for each input token, enabling powerful feature extraction, \textit{ii)} a span feature masking strategy to replace masking tokens with learnable tokens, effectively alleviating the impact of overfitting, and \textit{iii)} employ Sharpness-Aware Minimization (SAM) optimizer to ensure
that the model can converge towards flatter minima.

\paragraph{CNN feature extractor}
To make our pipeline simple, we adopt the widely-used ResNet-18~\cite{he2016deep} as our CNN feature extractor, with minor adjustments made to accommodate line-level handwritten text images. Specifically, we remove the final residual block and adjust the stride to produce features with enough information for character recognition while maintaining the two-dimensional nature of the task. More details about the modification as well as experiments of additional CNN feature extractors are provided in the supplementary material.



\paragraph{Span feature masking strategy} 
Our work draws inspiration from BERT \cite{devlin2018bert}, SpanBERT \cite{joshi2020spanbert} and MASS \cite{song2019mass}, which leverage the prediction of randomly masked words or tokens to learn expressive language representations. We have adapted this methodology to our specific task and observed the benefits of employing random feature masking. Furthermore, our intuition suggests that the feature extractor can capture a board receptive field. To enhance the model's comprehension of contextual information encompassing neighboring ink pixels, we propose expanding the masking range.

Precisely, the feature map after the CNN feature extractor is flattened to a sequence of tokens with dimensions $L \times C$, where $L$ represents the sequence length and $C$ represents the feature dimension. We randomly mask the span of tokens with a maximum span length $s$ (i.e., the number of interconnected tokens). In total $\tau L$ tokens are masked and replaced with a learnable token, where $\tau$ is a hyperparameter defining the mask ratio. More details of the span mask strategy are provided in the supplementary material.

\paragraph{Sharpness-Aware Minimization (SAM)} 
Sharpness-Aware Minimization (SAM), proposed by Foret et al. \cite{foret2020sharpness}, is an optimization method that enhances the generalization of deep neural networks (DNNs). It aims to find model parameters that reside in flat minima, ensuring a uniformly low loss across the model. Given our objective function $\mathcal{L}_{\text{CTC}}$ and the parameters of the DNN $\theta$, the SAM optimizer is designed to find $\theta$ such that:
\begin{equation} \label{eq:sam}
    \min_{\theta} \max_{\|\epsilon\|_2 \leq \rho} \mathcal{L}_{\text{CTC}}(\theta + \epsilon),
\end{equation}
where $\epsilon$ represents a perturbation vector, and $\rho$ is the size of the neighborhood within which the algorithm minimizes the sharpness of the loss function. The SAM algorithm functions by alternately identifying the worst-case perturbation $\epsilon$ that maximizes the loss within an $\ell_2$-norm ball of radius $\rho$, and then updating the DNN parameters $\theta$ to minimize this perturbed loss.

\paragraph{Implementation details}
We employ a ViT encoder with 4 layers. Each layer is with a dimension of 768 and 6 heads. The hidden dimension of MLP in the feed-forward network (FFN) is 3,072. Larger ViT models do not bring obvious gain. For our span mask strategy, we set the mask ratio to 0.4 and the span length to 8 in all datasets. An ablation study of the mask ratio and span length is provided in Section~\ref{sec:abl}. For all experiments, we use a batch size of 128 and optimize all our models with the AdamW \cite{loshchilov2017decoupled} optimizer for 100,000 iterations with a weight decay of 0.5. We perform a warm-up-cosine learning rate schedule with the max learning rate equal to 1e-3 and use 1,000 iterations for warm up. Trainings are performed on a single GPU RTX 4090 (24Gb) and in
the following experiments, models are trained for almost 16 hours. Similar to OrigamiNet~\cite{yousef2020origaminet}, we use the exponential moving average (EMA) method with a decay rate of 0.9999. For data augmentation, we fix the input image resolution to 512 x 64 and use random transformation, erosion, dilation, color jitter, and elastic distortion. We set the probability of using each data augmentation to 0.5, and they can be combined
with each other. 


\section{Experiments}
In this section, we evaluate the performance of our model for line-level recognition. Our experimental results demonstrate that our model achieves state-of-the-art results on the LAM~\cite{cascianelli2022lam} and IAM~\cite{marti2002iam} datasets. Moreover, our model competes well with other state-of-the-art models on the READ2016~\cite{sanchez2016icfhr2016} datasets. It is worth noting that our model achieves good performance without any pre-training or synthetic data and without relying on any pre/post-processing steps.

To further analyze the performance of our model, we conduct an ablation study by modifying the standard ViT \cite{dosovitskiy2020image} architecture. Specifically, we investigate the impact of different mask strategies and hyperparameters on the READ2016 dataset and examine how the SAM optimizer \cite{foret2020sharpness} affects our model's performance.

\subsection{Dataset and evaluation metrics}
We evaluated our model's performance on three commonly used datasets for handwritten text recognition: LAM \cite{cascianelli2022lam}, READ2016 \cite{sanchez2016icfhr2016}, and IAM \cite{marti2002iam}. Among these datasets, READ2016 and IAM are widely recognized as benchmarks for handwritten text recognition, while LAM is currently the largest available line-level handwritten text recognition dataset. The information about the datasets is provided in Table~\ref{tab:dataset}. Note that we report the performance on the test set with the model achieving the best performance on the validation sets. 

\paragraph{LAM~\cite{cascianelli2022lam}} 
The Ludovico Antonio Muratori (LAM) dataset is a massive handwritten text recognition dataset of Italian ancient manuscripts, which was edited by a single author over a span of 60 years. It consists of a total of 25,823 lines and has a lexicon of over 23,000 unique words. The dataset is split into 19,830 lines for training, 2,470 lines for validation, and 3,523 lines for testing, with a charset size of 89. The dataset was annotated at the line level, with each line's bounding box and diplomatic transcription provided. During the transcription process, stroke-out text, illegible words due to stains and scratches, and special symbols not representable in Unicode were replaced with the \# symbol. This is currently the largest line-level handwritten text recognition dataset available and could be an ideal choice for demonstrating the potential of our model.

\paragraph{READ2016~\cite{sanchez2016icfhr2016}} 
READ2016 was proposed in the ICFHR 2016 competition on handwritten text recognition. It comprises a subset of the Ratsprotokolle collection used in the READ project, with color images representing Early Modern German handwriting. The dataset provides segmentation at the page, paragraph, and line levels.  For line-level tasks, the dataset has a total of 8349 training images, 1040 validation images, and 1138 test images, with a character set size of 89.

\paragraph{IAM~\cite{marti2002iam}}
IAM is a well-known offline handwriting benchmark dataset for modern English. It comprises 1,539 scanned text pages of English texts extracted from the LOB corpus, which were handwritten by 657 different writers. The training set of IAM has 747 documents (6,482 lines), the validation set has 116 documents (976 lines), and the test set has 336 documents (2,915 lines). The IAM dataset consists of grayscale images of English handwriting with a resolution of 300 dpi. In this work, we utilized the line level with the commonly used split, as described in Table~\ref{tab:dataset}.
\begin{table}[!t]
    \begin{center}
    \scriptsize
			\begin{tabular}{|l|c|c|c|}
				\hline \hline 
				Method &\multicolumn{1}{|c}{Test CER}&\multicolumn{1}{c|}{Test WER} &\multicolumn{1}{c|}{Param.}\\
				\hline
            CNN + BLSTM$^{\star}$ \cite{puigcerver2017multidimensional} & 5.8 & 18.4 & 9.3M\\
            GFCN$^{\star}$ \cite{coquenet2020recurrence,cascianelli2022lam}& 5.2 & 18.5 & 1.4M\\
            CRNN$^{\star}$ \cite{shi2016end,cascianelli2022lam}& 3.8 & 12.9 & 18.2M\\
            OrigamiNet-12$^{\star}$ \cite{yousef2020origaminet,cascianelli2022lam} &3.1 & 11.2 &39.0M\\
            OrigamiNet-18$^{\star}$ \cite{yousef2020origaminet,cascianelli2022lam} &3.1 & 11.1 & 77.1M\\
            OrigamiNet-24$^{\star}$ \cite{yousef2020origaminet,cascianelli2022lam} & 3.0 & 11.0 & 115.3M\\	\hline
        \multicolumn{4}{|c|}{\textbf{Transformer-based models}} \\ \hline
            ViT \cite{dosovitskiy2020image} & 6.1 & 19.1 & 37M\\ 
            ViT + DropKey \cite{dosovitskiy2020image,li2023dropkey} & 5.7 & 16.5 & 37M\\ 
            DeiT \cite{touvron2021training} & 5.9 & 18.7 & 6M \\ 
            Transformer$^\S$$^{\star}$ \cite{kang2022pay,cascianelli2022lam}&10.2 & 22.0 & 54.7M\\
            TrOCR$^\S$$^{\star}$ \cite{li2023trocr,cascianelli2022lam} & 3.6 & 11.6 & 385.0M\\          
        \bf HTR-VT & \bf 2.8 & \bf 7.4 &53.5M  \\
        \hline \hline
        \multicolumn{4}{c}{\footnotesize{$^\S$ reports results using extra training data.}} \\
        \multicolumn{4}{c}{\footnotesize{$^{\star}$ indicates re-implementations by LAM \cite{cascianelli2022lam}.}}
    \end{tabular}
    \caption{\textbf{Comparison with  state-of-the-art  approaches on LAM~\cite{cascianelli2022lam} dataset (19,830 training samples).} We outperform all the competitive approaches with a clear margin. The improvement is more important for the transformer-based approaches.}
    \label{tab::lam}
    \end{center}
\end{table}

\begin{table}[!t]
    \begin{center}
    \scriptsize
		  \begin{tabular}{|l|c|c|c|}
				\hline \hline 
				   Method  & {Test CER}  &{Test WER} &\multicolumn{1}{|c|}{Param.} \\
				\hline
                    CNN + RNN \cite{sanchez2016icfhr2016} & 5.1 & 21.1 & - \\
				  CNN + BLSTM \cite{michael2019evaluating} & 4.7 & - & -\\
                    FCN \cite{coquenet2021span} & 4.6 & 21.1 & 19.2M \\
                    VAN \cite{coquenet2022end} & 4.1 & \bf 16.3 & 2.7M \\  \hline
			    \multicolumn{4}{|c|}{\textbf{Transformer-based models}} \\ \hline
                ViT \cite{dosovitskiy2020image} & 8.5 & 29.6 & 37M \\ 
                ViT + DropKey \cite{dosovitskiy2020image,li2023dropkey} & 8.1 & 26.4 & 37M \\ 
                DeiT \cite{touvron2021training} & 8.4 & 28.7 & 6M  \\ 	
                DAN \cite{coquenet2023dan}  & 4.1  & 17.6 & 7.6M \\ 
				\bf HTR-VT& \bf 3.9 & \bf 16.5 & 53.5M  \\
				\hline \hline
		    \end{tabular}
    \caption{\textbf{Comparison with state-of-the-art  approaches on READ2016~\cite{sanchez2016icfhr2016} dataset (8,349 training samples).} We achieve comparable performance.}
    \label{tab::read}
    \end{center}
\end{table}

\subsection{Comparison with state-of-the-art approaches}
\paragraph{Evaluation metrics}
We use Character Error Rate (CER) and Word Error Rate (WER) as performance measures. CER is calculated as the Levenshtein distance between two strings, which is the sum of character substitutions ($\mathrm{SUB_c}$), insertions ($\mathrm{INS_c}$), and deletions ($\mathrm{DEL_c}$) required to transform one string into the other, divided by the total number of characters in the ground truth ($\mathrm{GT_c}$). Formally, CER is given by:
\begin{equation}
    \mathrm{CER} = \frac{\mathrm{SUB_c} + \mathrm{INS_c} + \mathrm{DEL_c}}{\mathrm{GT_c}}.
\end{equation}
Similarly, $\mathrm{WER}$ is calculated as the sum of word substitutions ($\mathrm{SUB_w}$), insertions ($\mathrm{INS_w}$), and deletions ($\mathrm{DEL_w}$) needed to transform one string into another, divided by the total number of words in the ground truth ($\mathrm{GT_w}$). Mathematically, WER is expressed as:
\begin{equation}
    \mathrm{WER} = \frac{\mathrm{SUB_w} + \mathrm{INS_w} + \mathrm{DEL_w}}{\mathrm{GT_w}}
\end{equation}
\begin{table}[!t]
    \begin{center}
    \scriptsize
			\begin{tabular}{|l|c|c|c|}
				\hline \hline 
				Method &\multicolumn{1}{|c}{Test CER}&\multicolumn{1}{c|}{Test WER} &\multicolumn{1}{c|}{Param.}\\
				\hline
				GFCN \cite{coquenet2020recurrence} & 8.0 & 28.6 &1.4M \\
                GFCN$^\star$ \cite{coquenet2020recurrence,cascianelli2022lam} & 8.0 & 28.6& 1.4M \\
                CRNN$^\star$ \cite{shi2016end,cascianelli2022lam} & 7.8  & 27.8 & 18.2M \\
                CNN + BLSTM \cite{puigcerver2017multidimensional} & 8.3 & 24.9 & 9.3M  \\
                CNN + BLSTM$^\star$ \cite{puigcerver2017multidimensional,cascianelli2022lam} & 7.7 & 26.3 & 9.3M  \\
                OrigamiNet-12 \cite{yousef2020origaminet} &5.3 & - & 39.0M\\
                OrigamiNet-12$^\star$ \cite{yousef2020origaminet,cascianelli2022lam} &6.0 & 22.3 & 39.0M \\
                OrigamiNet-18 \cite{yousef2020origaminet} &4.8 & - & 77.1M \\
                OrigamiNet-18$^\star$ \cite{yousef2020origaminet,cascianelli2022lam} &6.6 & 24.2 & 77.1M \\
                OrigamiNet-24 \cite{yousef2020origaminet} &4.8 & - & 115.3M\\
                OrigamiNet-24$^\star$ \cite{yousef2020origaminet,cascianelli2022lam} &6.5 & 23.9 & 115.3M \\
                VAN \cite{coquenet2022end} & \bf 5.0 & \bf 16.3 & 2.7M  \\ \hline
				\multicolumn{4}{|c|}{\textbf{Transformer-based models}} \\ \hline

                ViT \cite{dosovitskiy2020image} & 32.4 & 68.5 & 37.0M \\ 
                ViT + DropKey \cite{dosovitskiy2020image,li2023dropkey} & 34.2 & 70.1 & 37.0M\\ 
                DeiT \cite{touvron2021training} & 32.0 & 68.4 & 6.0M \\ 
                    Transformer$^\S$ \cite{kang2022pay}& 4.7 & 15.5 & 54.7M  \\  
                    Transformer$^{\spadesuit}$ \cite{kang2022pay,cascianelli2022boosting}& 7.6 & 24.5 & 54.7M \\
                TrOCR$^\S$ \cite{li2023trocr} & 3.4 & - & 385.0M \\
				TrOCR$^\star$ \cite{li2023trocr,cascianelli2022lam} & 7.3 & 37.5 & 385.0M \\
				\bf HTR-VT& \bf 4.7 & \bf 14.9 &  53.5M \\
				\hline \hline
                \multicolumn{4}{c}{\scriptsize{ $^\S$ reports results using extra training data.} } \\
                \multicolumn{4}{c}{\scriptsize{ $^{\star}$ and $^{\spadesuit}$ indicate re-implementations by LAM \cite{cascianelli2022lam} and by \cite{cascianelli2022boosting}.}}
		      \end{tabular}
		\caption{\textbf{Comparison with state-of-the-art  approaches on the test set of IAM~\cite{marti2002iam} dataset (6,482 training samples).} Our approach exceeded the previous state-of-the-art model.}
		\label{tab::iam}
    \end{center}
\end{table}
We conducted a comparative study of current state-of-the-art methods on the LAM \cite{cascianelli2022lam}, READ2016 \cite{sanchez2016icfhr2016}, and IAM \cite{marti2002iam} datasets respectively. Our approach surpassed previous state-of-the-art models on the LAM \cite{cascianelli2022lam} and IAM \cite{marti2002iam} datasets and achieved comparable performance on the READ2016 \cite{sanchez2016icfhr2016} dataset. The results presented in Tables~\ref{tab::lam}, \ref{tab::read} and \ref{tab::iam} were achieved without the utilization of any external language models, such as n-grams or similar techniques. Specifically, on the LAM \cite{cascianelli2022lam} dataset, our method achieved a CER of 2.8 and a WER of 7.4, outperforming all models tested on this dataset. On the IAM \cite{marti2002iam} dataset, our approach exceeded the previous state-of-the-art model, VAN \cite{coquenet2022end}, with a CER improvement of 0.3 and a WER improvement of 1.4. On the READ2016 \cite{sanchez2016icfhr2016} dataset, our method reached a CER of 3.9, surpassing the state-of-the-art method VAN \cite{coquenet2022end} and DAN \cite{coquenet2023dan} by 0.2, and closely matching its WER.
\\
Furthermore, when compared to all transformer-based methods, our approach consistently led the field, except on the IAM dataset \cite{marti2002iam} where TrOCR \cite{li2023trocr} achieved a CER of 3.4. However, it is noteworthy that TrOCR \cite{li2023trocr} uses pre-trained CV and NLP models and a large-scale synthetic dataset, which is not publicly available, to pre-train their model. Transformer \cite{kang2022pay} also relies on a large amount of synthetic data for training. Despite this, our method still outperforms it. In addition, we also conduct a fair comparison to two recent works on data-efficient transformers: DeiT~\cite{touvron2021training} and DropKey~\cite{li2023dropkey}. We achieve clearly better performance than them on all three datasets. Training details of DeiT~\cite{touvron2021training} and DropKey~\cite{li2023dropkey} are provided in the supplementary material. These results demonstrate the data-efficiency of our proposed model.
\\
In summary, our research presents a competitive handwritten text recognition model that stands out against state-of-the-art methods, particularly on the LAM \cite{cascianelli2022lam} and IAM \cite{marti2002iam} datasets, and competes well on the READ2016 \cite{sanchez2016icfhr2016} dataset without resorting to any external language models, pre-training or synthetic data commonly used in the field. \\
For many years, the CNN + BLSTM paradigm has been the dominant approach in handwritten text recognition. However, our proposed method represents a significant shift in this trend, markedly enhancing the performance of transformer-based models. This breakthrough has the potential to steer the entire field of handwritten text recognition toward new and exciting directions.


\begin{table}[!t]
    \begin{center}
		\resizebox{\columnwidth}{!}{
			\begin{tabular}{|l|c|c|c|c|c|c|}
				\hline \hline 
				Methods &\multicolumn{2}{|c|}{LAM \cite{cascianelli2022lam}} &\multicolumn{2}{c|}{IAM \cite{marti2002iam}} &\multicolumn{2}{c|}{READ2016~\cite{sanchez2016icfhr2016}} \\ 
                    \hline
                        & \multicolumn{1}{c} {Val CER}     
                        & \multicolumn{1}{c|} {Val WER}  
                        & \multicolumn{1}{c} {Val CER}    
                        & \multicolumn{1}{c|} {Val WER}  
                        & \multicolumn{1}{c} {Val CER}  
                        & \multicolumn{1}{c|} {Val WER}   \\ 
			    
      ViT$^\star$  & 5.7  & 16.7 & 26.6 & 57.1 & 9.4 & 35.2 \\
                    Ours w/o. CNN extractor  & 5.5 & 15.7 & 20.7 & 53.5 & 8.9 & 33.7 \\  
				  Ours w/o. SAM  & 2.7 & 7.4 & 3.4 & 11.2 & 4.8 & 20.1 \\
                    Ours w/o. Span Mask  & 2.9 & 7.8 & 3.7 & 12.1  & 5.1 & 21.9 \\
				  \bf Ours  & \bf 2.6 &  \bf 6.9 & \bf 3.3 & \bf 10.8  &  \bf 4.5 & \bf 19.4 \\
		          \hline \hline
                \multicolumn{7}{c}{$^\star$ ViT is equivalent to our approach without CNN extractor nor Span Mask.}
		    \end{tabular}}
      \caption{{\bf Ablation study of our approach on LAM \cite{cascianelli2022lam}, IAM \cite{marti2002iam} and READ2016 \cite{sanchez2016icfhr2016} datasets.} We reported the performance of the standard ViT and studied the effect of our architecture without CNN feature extractor, SAM, Span Masking, respectively, on the results.}
		\label{tab::w/oablation}
  \end{center}
\end{table}
\begin{table}[!t]
    \begin{center}
		\resizebox{\columnwidth}{!}{
			\begin{tabular}{|cc|c|c|c|c|c|c|c|c|}
				\hline \hline 
				&&\multicolumn{4}{c|}{IAM \cite{marti2002iam}} &\multicolumn{4}{c|}{READ2016~\cite{sanchez2016icfhr2016}} \\ 
                    \hline
                       \multicolumn{1}{|c} {Layers} & \multicolumn{1}{c|} {Heads} & \multicolumn{1}{c} {Val CER} 
                       & \multicolumn{1}{c|} {Val WER}
                        & \multicolumn{1}{c} {Test CER}
                        & \multicolumn{1}{c|} {Test WER} 
                        & \multicolumn{1}{c} {Val CER}
                        & \multicolumn{1}{c|} {Val WER}
                        & \multicolumn{1}{c} {Test CER} 
                        & \multicolumn{1}{c|} {Test WER}
                        \\ 
                   8 & 6  &  3.6 & 11.8 & 5.2  & 16.2 & 4.8 & 20.1 & 4.2 & 17.6 \\  
				\bf 4& 6 & 3.3   & 10.8 & 4.7 & 14.9 & 4.5 & 19.4 & 3.9  & 16.5 \\
                    2& 6 & 3.5  & 11.4   & 5.1 & 16.1 & 4.3 & 18.8 & 3.9 & 16.7 \\
				  1& 6  & 4.1 & 13.6 & 6.0 & 18.9 & 5.0  & 21.4 & 4.8 & 20.0\\
		          \hline 
            \multicolumn{1}{|c}{Layers} & \multicolumn{1}{c|}{Heads} & \multicolumn{1}{c} {Val CER} 
            & \multicolumn{1}{c|} {Val WER}
                        & \multicolumn{1}{c} {Test CER}
                        & \multicolumn{1}{c|} {Test WER} 
                        & \multicolumn{1}{c} {Val CER} 
                        & \multicolumn{1}{c|} {Val WER} 
                        & \multicolumn{1}{c} {Test CER} 
                        & \multicolumn{1}{c|} {Test WER} \\ 
                   4 & 8  & 3.5  & 11.3 &  4.9 & 15.6 & 4.6 & 20.0 & 4.1 &17.4  \\  
				4 &  \bf 6 & 3.3   & 10.8 & 4.7 & 14.9 &4.5 & 19.4 & 3.9  & 16.5 \\  
                    4& 4 & 3.3  & 10.9 & 4.7 & 14.8 & 4.4 & 18.8& 4.1 & 17.6\\
                    4& 2 & 3.5 &  11.4 & 5.1 & 16.0 & 4.5 & 19.5 & 4.0 & 17.7 \\ 
                \hline \hline 
		    \end{tabular}}
      \caption{{\bf Ablation study of more hyperparameters on IAM \cite{marti2002iam} and READ2016 \cite{sanchez2016icfhr2016} datasets.} We studied the effect of different transformer encoder layers and attention heads on the results.}
		\label{tab::parameters}
  \end{center}
\end{table}

\subsection{Ablation studies and visualization analysis}
In this section, we delve into two core areas of our study: ablation studies and visualization analysis. The ablation studies are comprehensive, examining the impact of key building blocks within our model and exploring the influence of decoder and critical hyperparameters. These include the masking ratio and span length, as well as the number of transformer encoder and decoder layers and attention heads. The visualization analysis grants us deeper insights into the effectiveness of our span mask strategy.
Additionally, we present several qualitative results that showcase the effectiveness of our model.

We hope that our research can serve as a solid basis that can be readily and swiftly used by future researchers. For this reason, we have intentionally refrained from incorporating intricate and opaque components into our model, which could pose difficulties in explanation.
\paragraph{Effect of CNN feature extractor}
We achieved relatively good results on LAM \cite{cascianelli2022lam} and READ2016 \cite{sanchez2016icfhr2016} datasets using only the standard ViT encoder. This encouraged us to consider the ViT architecture as a promising approach for handwritten text recognition tasks. However, we observed that training with the ViT encoder alone resulted in unstable performance and slow convergence speed on the IAM dataset \cite{marti2002iam}, making it difficult to compete with CNN-based models. To improve performance, we introduced a CNN-based feature extractor before the ViT encoder to combine the transformer's global feature extraction capabilities with the CNN's ability to extract local features via a strong inductive bias. Our experiments showed that this modification significantly improved the model's performance and convergence speed.

\begin{figure}[!t]
\centering
\includegraphics[width=0.48\textwidth]{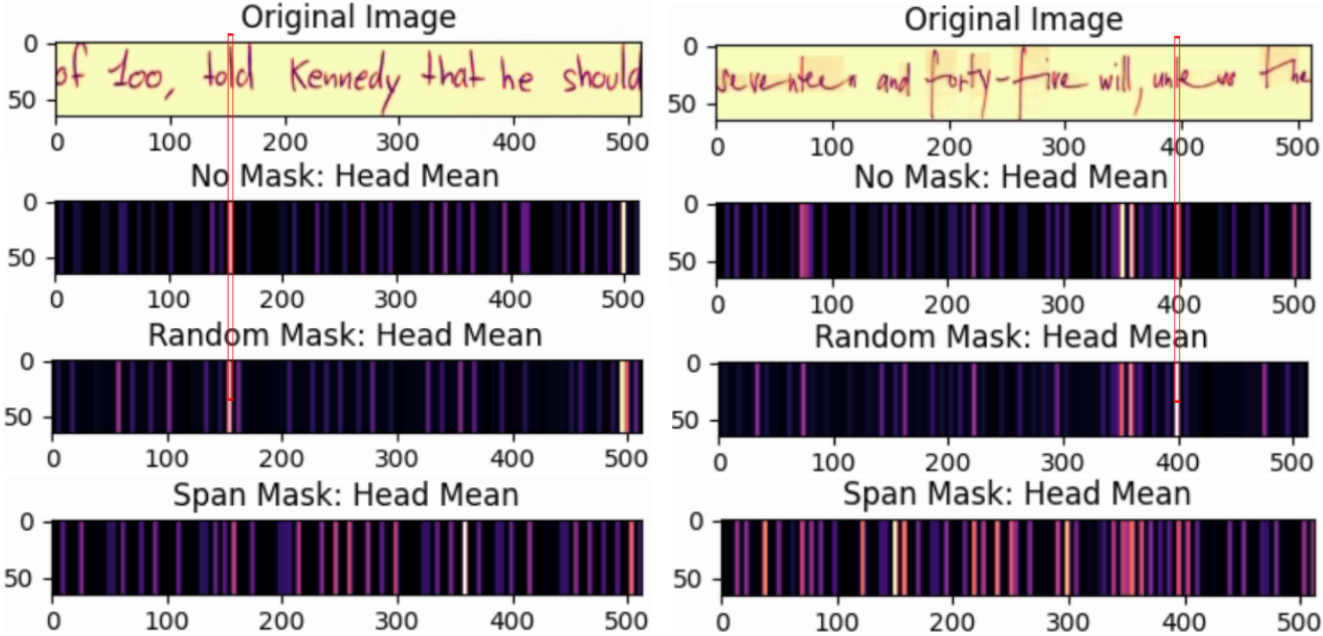}
    \caption{{\bf Visualization of attention maps with different masking strategies on IAM dataset.} In the original image, we highlight the region corresponding to the token of interest with a red bounding box and average the attention across all heads. We observe that when no masking or random masking strategy is employed, each token focuses solely on its own information, as indicated by the illuminated regions in the image. However, when we apply the span masking strategy, a noticeable shift occurs, allowing the token to attend to a broader range of information.}
\label{fig:attention map}
\end{figure}

\paragraph{Effect of employing Sharpness-Aware Minimization(SAM) \cite{foret2020sharpness} optimizer}
We found that convergence to a flatter minimum can mitigate overfitting in Handwritten Text Recognition (HTR) models. To facilitate this, we utilized the Sharpness-Aware Minimization (SAM) optimizer, which is straightforward to apply, for locating these flatter minima. Our experimental results show that validation CER and WER on READ2016 increased from 4.8 to 4.5 and from 20.1 to 19.4 with SAM \cite{foret2020sharpness}optimizer, indicating that it has a significant impact on HTR tasks.

\paragraph{Effect of span feature masking}
When labeled data is limited, overfitting can become problematic for transformer-based models. To address this issue, we proposed a new feature masking strategy to reduce overfitting and improve model performance as described in section \ref{sec:ours}. As shown in Table \ref{tab::w/oablation}, the span feature masking provides consistent and clear improvement across all the datasets.


\paragraph{Impact of different hyperparameters}
We investigate the impact of different transformer encoder layers and attention heads. The results are illustrated in Table \ref{tab::parameters}. On IAM dataset, taking the number of layers to 4 and attention heads of 6 achieved the best validation CER and WER. To maintain consistency, we employed this set of parameters across all our experiments.
We also investigate the impact of different masking strategies. The results are illustrated in Table \ref{tab::maskratio}.
We can see masking tokens (\textit{`Span Length = 1'}) or span feature masking strategy (\textit{`Span Length > 1'}) improve the performance for most cases.  
Span feature masking performs better than random masking tokens and masking none of the tokens. 
For hyperparameters, taking the masking ratio of 0.4 and a span length of 8 is optimal, which is used for all our experiments. However, larger span lengths (16) reduce the performances, possibly due to the inability to learn context-related information.

\paragraph{Impact of transformer decoder}
Similar to TrOCR \cite{li2023trocr}, we employ a standard transformer decoder and utilize beam search to produce the final output. We utilized our optimal encoder as the baseline to systematically investigate the impact of the decoder on the overall model performance. The increased number of parameters from adding a decoder constrained us to use a batch size of 64 to maintain consistency across all ablations. Our experiments with decoders of varying layer counts, as shown in Table \ref{tab::decoder ablations}, demonstrated that incorporating a transformer decoder did not facilitate better convergence nor prevent overfitting.
\begin{table}[!t]
    \begin{center}
        \scriptsize
			\begin{tabular}{|c|c|c|c|}
                \hline \hline 
				Mask Ratio  & Span Length & \multicolumn{1}{c|} {Val CER}    & \multicolumn{1}{c|} {Val WER}   \\
				\hline
                    0.0 & None$^\star$ & 5.1 & 21.9 \\ \hline 
                    0.2 & \multirow{3}{*}{1$^\S$} & 4.9 & 20.6 \\
                    0.4 &  &4.8 & 20.0\\
                    0.6 &  &5.1 & 21.8 \\ \hline 
                    0.2 & \multirow{3}{*}{4} & 4.6 & 19.8 \\
                    0.4 &  & 4.7 & 20.1\\
                    0.6 &  & 4.9 & 20.7 \\ \hline 
                    0.2 & \multirow{3}{*}{\bf 8} & 4.6 & 19.7 \\
                    \bf 0.4 &  & \bf 4.5 & \bf 19.4 \\
                    0.6 &  & 4.9 & 20.9 \\ \hline 
                    0.2 & \multirow{3}{*}{16} & 5.0 & 21.5 \\
                    0.4 &  & 5.2 & 22.6 \\
                    0.6 &  & 5.3 & 23.1 \\
				\hline \hline
                    
                    \multicolumn{4}{c}{\scriptsize{ $^\star$ indicates our approach without any masking strategy.} } \\
                \multicolumn{4}{c}{\scriptsize{ $^\S$ is equivalent to standard random masking.}}
		      \end{tabular}
		\caption{{\bf Ablation study on the masking strategy on READ2016 \cite{sanchez2016icfhr2016} dataset.} We studied the effect of different mask ratios.}
  
		\label{tab::maskratio}
  \end{center}
 \end{table}

\paragraph{Visualization of attention maps}
In our study, we examine the variations in attention maps when different masking strategies are employed in Figure \ref{fig:attention map}. 
We averaged the attention across all heads to generate the attention maps displayed. The detailed explanations are as follows:  \\
Firstly, our image size is fixed at 64 x 512, which, after patch embedding, transforms into a shape of 1 x 128, viewed as 128 tokens represented by 128 vertical stripes in the figure. The tokens selected for visualization correspond to the areas enclosed in red boxes in the original image. In the left image, the letter "o" is highlighted, while in the right image, it is the letter "l". According to the principle of self-attention, our selected token should pay more attention to other tokens with higher similarity, which is represented as lighter colors in the attention map. In both no-mask and random-mask scenarios, we can observe that in the left image, the letter "o" in "nvasion" and "bodies" is highlighted, and in the right image, the two "l" letters in "will" are illuminated. This indicates that under no mask and random mask conditions, attention is mainly focused on the token itself.
\\
However, a significant change is observed in the span masking scenario. More areas are noticed, indicating that when using span masking, tokens are able to "attend to a broader range of information". This highlights the effectiveness of span masking in enabling tokens to capture more contextual information. The improved contextual awareness provided by span masking facilitates a more comprehensive understanding of the text, which is vital for accurate recognition in handwritten text recognition tasks. \\
The more examples of the attention maps are provided in the supplementary material.

\paragraph{Comparison of training time}
Few methods mention the total time required to complete their training, yet this is extremely important for this task. Most approaches that rely on pre-training or additional data consume significantly expensive computational resources. We compared our method with CNN-based approaches GFCN \cite{coquenet2020recurrence}, VAN \cite{coquenet2022end} and OrigamiNet-24 \cite{yousef2020origaminet} in Table \ref{tab::training time}. It is important to highlight that the VAN method did not resize images to a fixed resolution but instead used the original image pixels from datasets such as IAM \cite{marti2002iam}. Similarly, GFCN mentioned that the experiments for the IAM dataset with an image height of 128px, preserving the original width. This resolution is much larger than the fixed resolution we used, which is 512x64. OrigamiNet also used a fixed resolution of 600x32, and our approach of using a fixed resolution follows OrigamiNet. As shown in Table \ref{tab::training time}, our proposed transformer-based method remains competitive in terms of training time.

\begin{figure}[!t]
\centering
\includegraphics[width=\columnwidth]{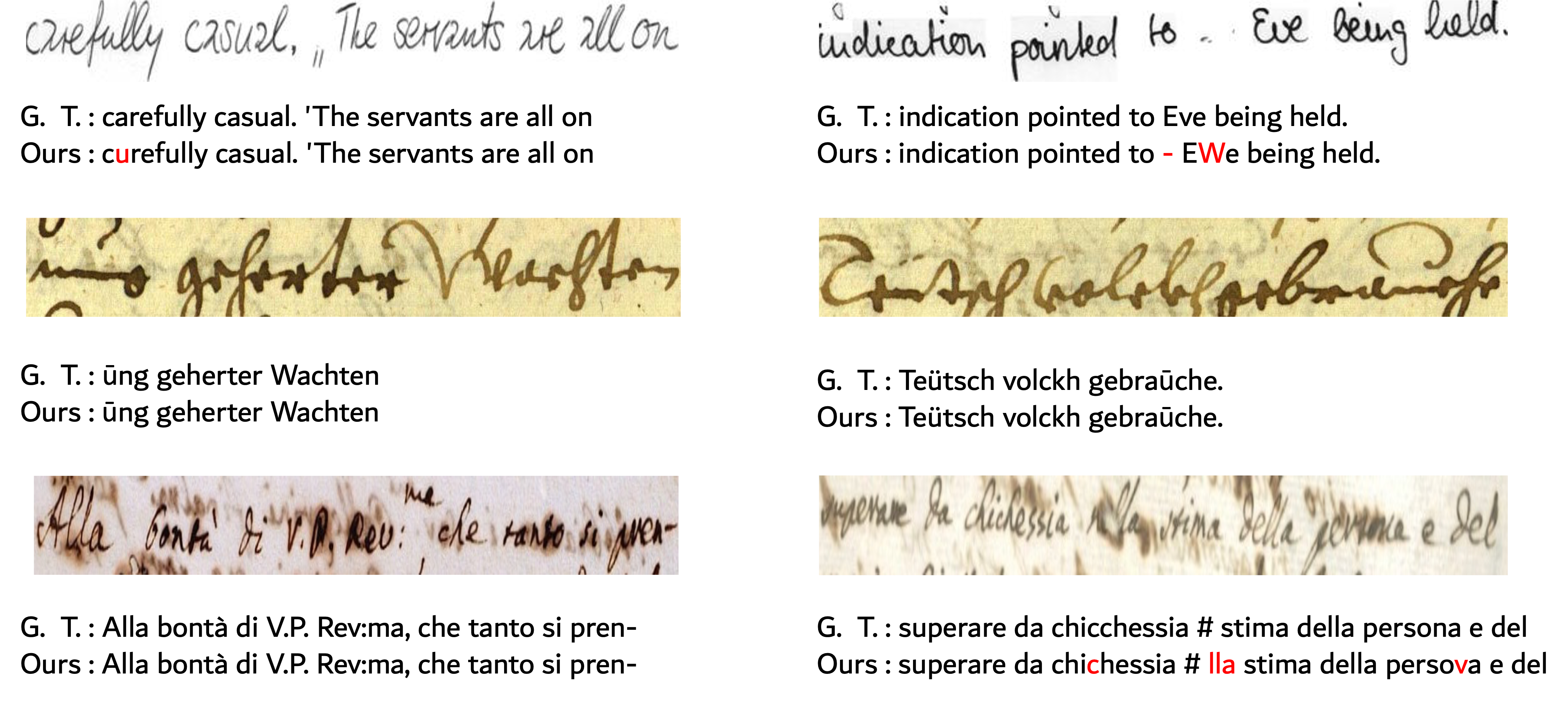}
\caption{Results on example lines from the IAM \cite{marti2002iam} (First row), READ2016 \cite{sanchez2016icfhr2016} (Second row) and LAM \cite{cascianelli2022lam} (Third row) of the best performing model. }
\label{fig:results}
\end{figure}

\paragraph{Qualitative results} 
We provide visual results in Figure~\ref{fig:results} for IAM~\cite{marti2002iam} (First row), READ2016~\cite{sanchez2016icfhr2016} (Second row) and LAM~\cite{cascianelli2022lam} (Third row). From this, one can recognize the task is challenging, as the visual content present in the text line is not very visible and the background is quite noisy. However, our approach can still produce reasonable predictions on these examples. It is worth noting that in the final image, the ground truth label was annotated incorrectly. Despite this error, our proposed model was still able to accurately recognize the correct handwritten text from the original image, which demonstrates the robustness and effectiveness of the proposed approach.

\begin{table}[!t]
    \begin{center}
		\resizebox{0.8\columnwidth}{!}{
			\begin{tabular}{|c|c|c|c|c|c|}
				\hline \hline 
				\multicolumn{1}{|c|}{}&\multicolumn{5}{c|}{IAM \cite{marti2002iam}}  \\ 
                    \hline
                       \multicolumn{1}{|c|}{Decoder Layers} & \multicolumn{1}{c} {Val CER }   & \multicolumn{1}{c|} {Val WER}
                        & \multicolumn{1}{c} {Test CER}   & \multicolumn{1}{c|} {Test WER } &\multicolumn{1}{|c|}{Param.} \\
                    \hline
                    0 &3.3 &10.8&4.7 &14.9 & 53.5M\\
                    \hline
                   8  &  - & - & -  & - &129.2M \\  
				 4 & 5.0 & 15.3 & 7.7 & 21.3 & 91.4M \\
                    2 & 5.0  & 15.1 & 7.8 & 21.1 & 72.5M \\
				  1  & 5.3 & 15.7 & 8.1 & 21.6 & 63.0M \\
                \hline \hline 
		    \end{tabular}}
      \caption{{\bf Ablation study of adding decoder and training more iterations on IAM \cite{marti2002iam} dataset.} We studied the effect of different add transformer decoder layers and training iterations on the results. When the number of decoder layers is 8, the model is hard to converge.}
		\label{tab::decoder ablations}
  \end{center}
\end{table}
\begin{table}[!t]
    \begin{center}

		\resizebox{\columnwidth}{!}{
			\begin{tabular}{|c|c|c|c|c|c|}
				\hline \hline 
                       \multicolumn{1}{|c}{Architecture} &\multicolumn{1}{|c|}{1k iters} & {Total number of epochs / iters}& {Training time} & \multicolumn{1}{c|} {Param.} & \multicolumn{1}{c|} {Input size(H x W)} \\
                    \hline
                    GFCN \cite{coquenet2020recurrence}& 543 s &186 (Early stop) / 75.6k iters & 11.4 h  &1.4M & 128 x original W\\
VAN \cite{coquenet2022end} & 420 s & 2100 (Early stop) / 850.4k iters & 99.3 h  & 2.7M  & original H x W \\ 
OrigamiNet-24 \cite{yousef2020origaminet} & 476 s & 100k iters & 13.2 h & 115.3M & 32 x 600\\
				HTR-VT & 586 s \bf& 100k iters& 16.3 h &  53.5M & 64 x 512 \\
                \hline \hline 
		    \end{tabular}}\\
      \footnotesize{Note that the reported training times are approximate. }
      \caption{{\bf Comparison of the training times across different methods.}We have re-implemented the above methods to compare training times, using the same batch size of 64.}
		\label{tab::training time}
  \end{center}
\end{table}

\label{sec:abl}

\section{Discussion}
Although our approach has made notable strides in transformer-based line-level recognition, there is still room for improvement in our current method. The significance of data augmentation for handwriting recognition cannot be overstated, and we have observed that certain data augmentation methods previously utilized in HTR may have adverse effects. Investigating new types of data augmentation specifically tailored for handwriting is a potential direction. Furthermore, delving deeper into mask strategies represents an intriguing avenue for exploration; learnable mask strategies adapted for handwriting could prove more beneficial. Lastly, expanding from line-level to paragraph-level or page-level recognition will be the focus of our future research.

\section{Conclusion}
In this work, we have presented a simple and data-efficient approach for handwritten text recognition. With minimal modifications to the ViT architecture, we have successfully developed a ViT-like model that surpasses state-of-the-art performance without requiring pre-training or additional data. Notably, our experiments highlight the remarkable data efficiency of our model compared to ViT and DeiT, while preserving its superior generalizability even in scenarios with vast amounts of available data. These findings provide a promising direction for improving the performance of handwritten text recognition, particularly in limited data scale settings. 




{\small
\bibliographystyle{elsarticle-num.bst}
\bibliography{main}
}

\clearpage
\newpage

\appendix
\vspace*{1em}{\centering\Large\bf%
Appendix
\vspace*{1.5em}}

This appendix contains the following sections: 
\begin{itemize}[itemsep=0.1em]
	\item ~\ref{sec:viusal_res}: visual results on handwritten text recognition datasets as mentioned in {\bf Introduction} .  
	\item ~\ref{sec:res_vgg_ablation}: more details about our Convolutional Neural Network(CNN) backbone and ablations on both ResNet and VGG. This experiment was mentioned in {\bf Section 3.2} of our paper.
        \item ~\ref{sec:span_mask}: more details about the span mask strategy. This experiment was mentioned in {\bf Section 3.2} of our paper.
        \item ~\ref{sec:deit_dropkey}: training details about DeiT and DropKey. This experiment was mentioned in {\bf Section 4.2} of our paper.
        \item ~\ref{sec:atten_vis}: some visualization results of attention maps as mentioned in {\bf Section 4.3} of our paper .
\end{itemize}

\begin{table}[!htb]
    \begin{center}
		\resizebox{0.6\columnwidth}{!}{
			\begin{tabular}{|l|c|c|}

				\hline \hline 
				Methods &\multicolumn{2}{c|}{IAM \cite{marti2002iam} } \\ 
                    \hline
                        &\multicolumn{1}{c} {Test CER}   &\multicolumn{1}{c|} {Test WER}
                        \\ 
                        
				  \bf ResNet-18   & 4.7  & 14.9  \\
				ResNet-50  & 4.9  & 15.6   \\
                    VGG-16   & 7.2 & 22.1  \\
		          \hline \hline
		    \end{tabular}}
      \vspace{3mm}
        \caption{{\bf Ablation study on using various CNN backbones on IAM \cite{marti2002iam} dataset.} We reported the performance of the our approach using different backbones and studied the effect of them. We use {\bf ResNet-18 } as our final solution. }
		\label{tab:backbone_ablation}
	\end{center}
\end{table}

\section{Visual results on the IAM \cite{marti2002iam}, READ2016 \cite{sanchez2016icfhr2016} and LAM \cite{cascianelli2022lam} datasets.}
\label{sec:viusal_res}
We show our handwritten text recognition method's visual results on the IAM \cite{marti2002iam}, READ2016 \cite{sanchez2016icfhr2016}, and LAM \cite{cascianelli2022lam} datasets.

\begin{figure}[!t]
    \centering
    \includegraphics[width=0.5\textwidth]{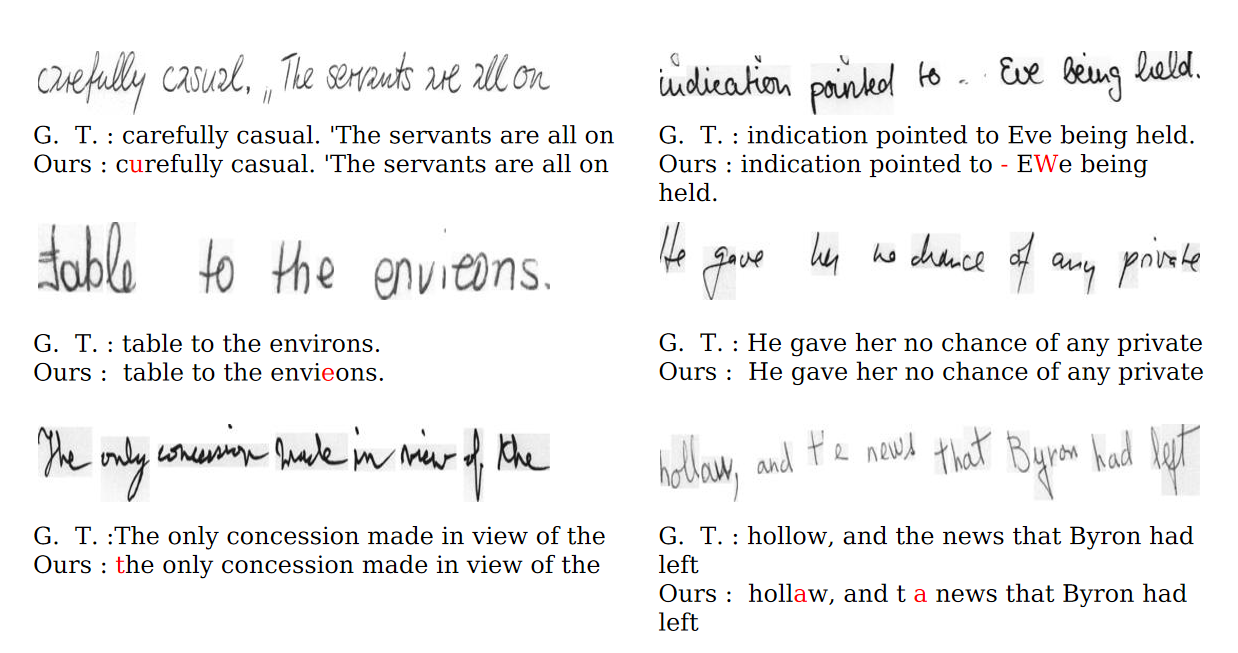}
    \caption{Visual results on IAM \cite{marti2002iam}}
    \label{fig:IAM}
    \includegraphics[width=0.5\textwidth]{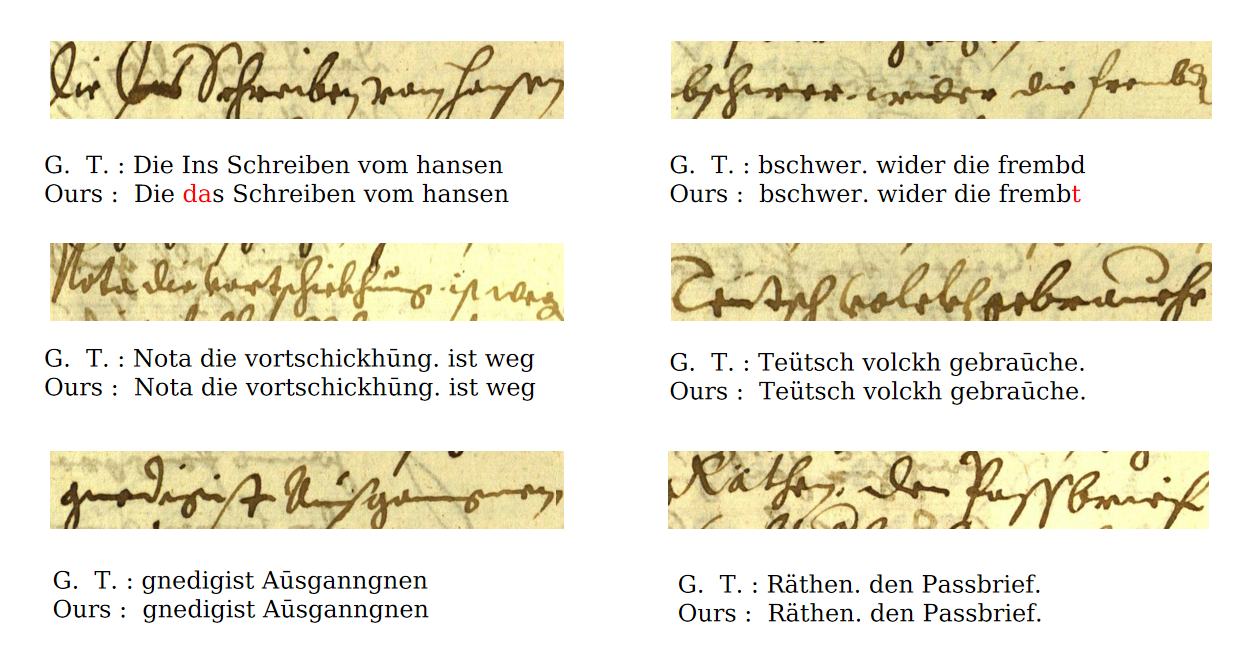}
    \caption{Visual results on READ2016 \cite{sanchez2016icfhr2016}}
    \label{fig:READ}
    \includegraphics[width=0.5\textwidth]{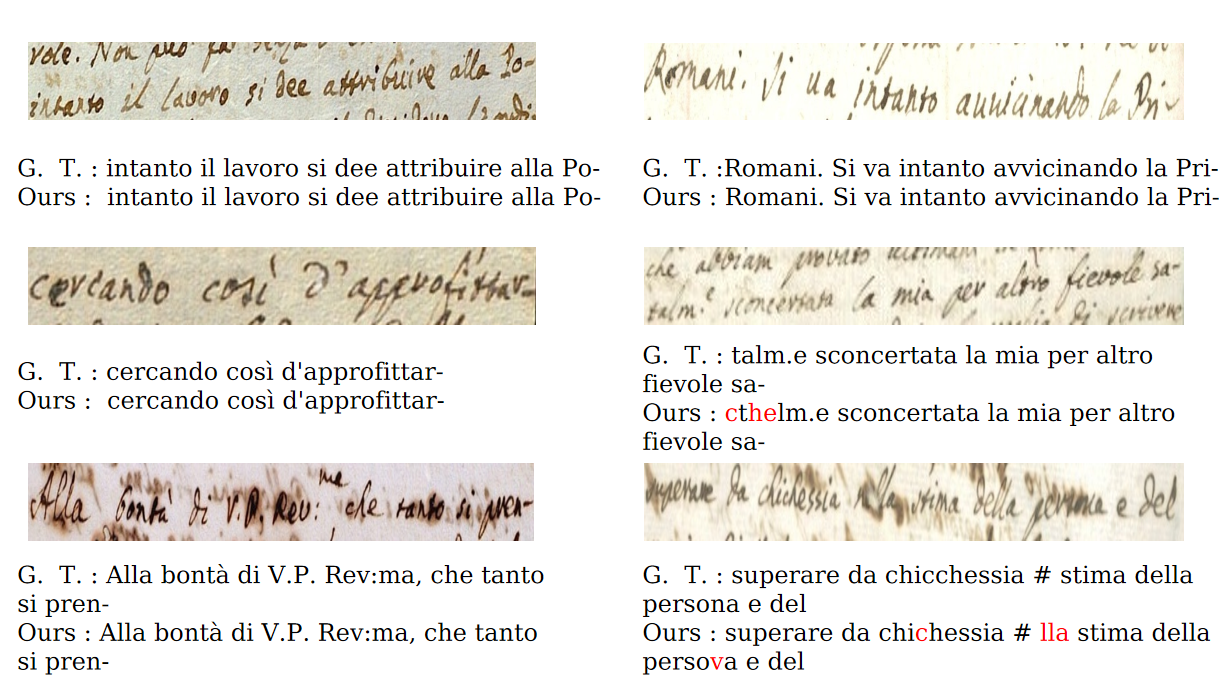}
    \caption{Visual results on LAM \cite{cascianelli2022lam}}
    \label{fig:LAM}
\end{figure}

IAM \cite{marti2002iam}  is a well-known offline handwriting benchmark dataset containing 6 482 images for training, 976 images for validation, and 2 915 images for testing. The image is in grayscale and the font has ligatures and some missing parts. Visual results are provided in Figure~\ref{fig:IAM}.

READ2016 \cite{sanchez2016icfhr2016} consists of 8 349 train, 1 040 validation, and 1 138 test images. The image contains a noisy background with some blurry fonts. Visual results are
provided in Figure~\ref{fig:READ}.

LAM \cite{cascianelli2022lam} is currently the largest line-level handwritten text recognition dataset that contains 19 830 lines for training, 2 470 lines for validation, and 3 523 lines for testing. The image contains fonts with stains, some lines of text are skewed and include both upper and lower characters. Visual results are provided in Figure~\ref{fig:LAM}.

\section{CNN Backbones Ablation}
\label{sec:res_vgg_ablation}
In this study, we investigated the impact of different CNN backbones on the overall model performance. We chose the most fundamental ResNet \cite{he2016deep}and VGG \cite{simonyan2014very}architectures as our CNN backbones, consistent with the simple and easy-to-implement principles outlined in our paper. The performance of the proposed method is observed to be robust across various backbones. Particularly, ResNet-18 exhibits superior performance compared to other backbones.

\section{Span mask strategy}
\label{sec:span_mask}
The details of our implementation of the span mask strategy are as follows: 
To achieve the designated mask ratio $R$ (e.g., 0.4 of $L$), we adopt an iterative process of sampling spans. In each iteration, we start by defining a maximum span length $l$ (i.e., the number of interconnected tokens), and then randomly select the starting point for each span. Noting that the maximum span length is fixed. This means that the length of the sampled masked segments remains the same for each iteration.
\begin{figure}[!t]
		\centering

        \includegraphics[width=0.5\textwidth]{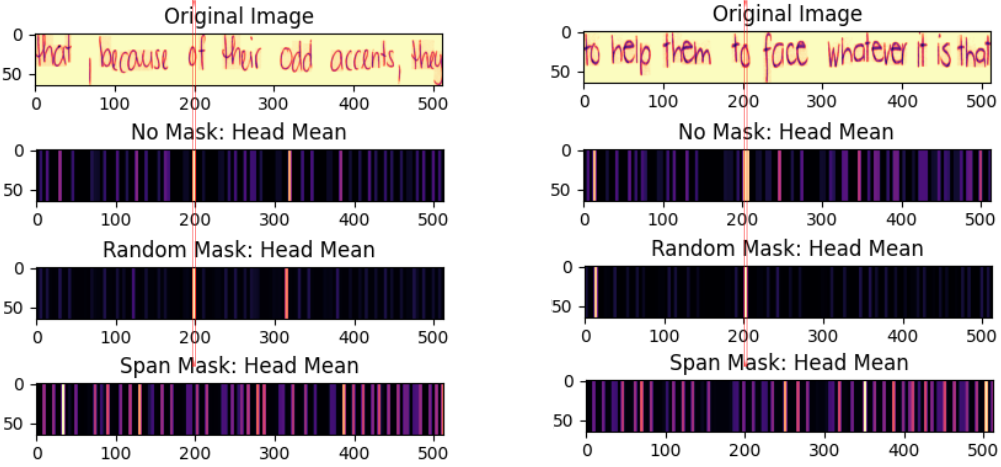}
        \includegraphics[width=0.5\textwidth]{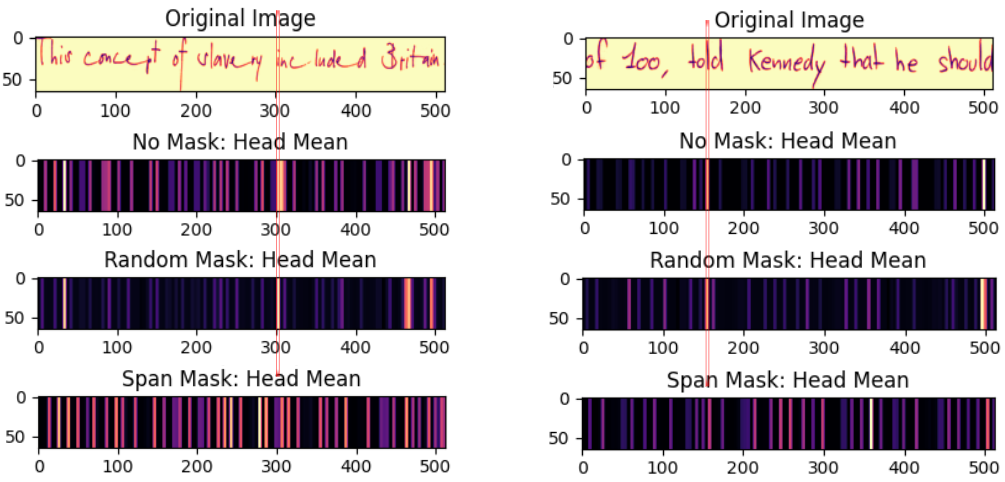}
        \caption{Visualization of attention maps}
        \label{fig:1213}
	\end{figure}


\section{Training details about DeiT \cite{touvron2021training}and DropKey \cite{li2023dropkey}}
We implemented it completely following the steps in DropKey \cite{li2023dropkey}, moving dropout operations ahead of attention matrix calculation and setting the Key as the dropout unit, yielding a dropout-before-softmax scheme. And We set the drop ratio to 0.1. In DeiT \cite{touvron2021training}, each layer has a dimension of 768 and 6 heads as used in our approach. At the same time, we implemented DeiT with no distillation.
\label{sec:deit_dropkey}

\section{Visualization results of attention maps}
In Figure \ref{fig:1213}, we present an extensive set of attention map visualizations that offer valuable insights into the model's behavior. We demarcate the region of interest in the original image corresponding to the token under scrutiny using a red bounding box. It is evident that when employing no mask and random mask strategies, the attention is highly localized, illuminating only the regions that correspond to the annotated characters in the original image. For instance, in the first visualization, the selected token corresponds to the letter 'o' in the word 'of,' and the attention map distinctly highlights this specific region. This suggests that, in these scenarios, each token is predominantly self-attentive.
Conversely, when utilizing a span mask strategy, there is a conspicuous expansion in the illuminated regions, indicating that the token now engages with a substantially broader contextual landscape.
\label{sec:atten_vis}


\end{document}